\documentclass[sn-mathphys-num]{sn-jnl}
\usepackage{graphicx}
\usepackage{multirow}
\usepackage{amsmath,amssymb,amsfonts}
\usepackage{amsthm}
\usepackage{mathrsfs}
\usepackage[title]{appendix}
\usepackage{xcolor}
\usepackage{textcomp}
\usepackage{manyfoot}
\usepackage{booktabs}
\usepackage{algorithm}
\usepackage{algorithmicx}
\usepackage{algpseudocode}
\usepackage{listings}
\usepackage{tabularx}
\usepackage{svg}
\usepackage{subcaption} 
\usepackage[utf8]{inputenc} 
\usepackage[T1]{fontenc}  
\usepackage{hyperref}

\newcommand{\myorcid}[1]{
    \href{https://orcid.org/#1}{
        \includegraphics[width=12pt]{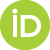}%
    }
}

\begin{document}

\title{Computer vision training dataset generation for robotic environments using Gaussian splatting}

\author*[1]{\fnm{Patryk} \sur{Niżeniec}\myorcid{0009-0001-1859-9299}}\email{pnizeniec@umk.pl}

\author[1]{\fnm{Marcin} \sur{Iwanowski}\myorcid{0000-0001-8347-1112}}\email{iwanowski@fizyka.umk.pl}

\affil[1]{\orgname{Institute of Engineering and Technology, Faculty of Physics, Astronomy and Informatics, Nicolaus Copernicus University in Toruń}, \orgaddress{\postcode{87-100}, \city{Toruń}, \country{Poland}}}

\abstract{This paper introduces a novel pipeline for generating large-scale, highly realistic, and automatically labeled datasets for computer vision tasks in robotic environments. Our approach addresses the critical challenges of the domain gap between synthetic and real-world imagery and the time-consuming bottleneck of manual annotation. We leverage 3D Gaussian Splatting (3DGS) to create photorealistic representations of the operational environment and objects. These assets are then used in a game engine where physics simulations create natural arrangements. A novel, two-pass rendering technique combines the realism of splats with a shadow map generated from proxy meshes. This map is then algorithmically composited with the image to add both physically plausible shadows and subtle highlights, significantly enhancing realism. Pixel-perfect segmentation masks are generated automatically and formatted for direct use with object detection models like YOLO. Our experiments show that a hybrid training strategy, combining a small set of real images with a large volume of our synthetic data, yields the best detection and segmentation performance, confirming this as an optimal strategy for efficiently achieving robust and accurate models.}


\maketitle

\section{Introduction}

Object detection and semantic segmentation play a crucial role in modern computer vision. It allows fast and accurate identification of position, size, and type of objects present within the digital image, which makes it applicable to many computer-vision-based engineering tasks~\cite{liu2020deepSurvey,ren2024objectsegmentation}, including robotics~\cite{manakitsa2024review}. The rapid advancement of robotics and industrial automation heavily relies on robust and accurate perception systems. Object detection and segmentation are fundamental tasks that enable robots to understand and interact with their environment. Modern solutions, particularly those based on deep learning architectures like YOLO (You Only Look Once)~\cite{kotthapalli2025yolov1v11, ramos2025decadeyolo}, have achieved remarkable performance. However, their effectiveness is directly proportional to the quality and quantity of the training data.

Object detection and semantic segmentation tasks belong to supervised learning, meaning they require annotated training data -- images with appropriate labeling. Creating large-scale, precisely annotated datasets is a significant bottleneck in developing robotic vision systems. Manual labeling is a labor-intensive, time-consuming, and error-prone process. Synthetic data generation emerges as a promising alternative, offering the potential to create vast, diverse, and perfectly labeled datasets at a fraction of the cost. However, a significant challenge with synthetic data is the "domain gap" -- the discrepancy between the appearance of synthetic images and their real-world counterparts, which can degrade model performance.

This paper introduces a novel pipeline for generating highly realistic, automatically labeled images -- training data for robotic environments. Our approach leverages 3D Gaussian Splatting (3DGS), a state-of-the-art neural rendering technique, to create photorealistic representations of operational environment and individual objects. To further bridge the domain gap, we introduce a unique method for simulating and compositing physically-based shadows, a crucial visual cue often missing in other synthetic data generation methods.

To validate the effectiveness of our proposed pipeline, we conduct a series of experiments. We train several YOLO11 models of varying sizes on three distinct datasets: one consisting of real-world images , a second generated entirely by our pipeline , and a hybrid set that combines both data types. The performance of each model is then evaluated on a consistent, held-out test set of real images. This comparison is quantified using standard mean Average Precision (mAP) metrics for both object detection and segmentation.

Our main contributions are:

\begin{enumerate} 
\item  a complete, end-to-end pipeline for generating synthetic datasets using Gaussian Splatting, 
\item a novel technique for isolating object splats and integrating them with physics-enabled meshes, and 
\item a hybrid rendering approach that combines photorealistic splats with simulated shadows to enhance realism. 
\end{enumerate} 

We demonstrate through experiments that a model trained on a hybrid dataset, combining a small number of real images with our synthetic data, outperforms models trained on either real or synthetic data alone.

This paper consists of 6 sections. Section 2 summarizes related works. In section 3, preliminary issues are introduced. Section 4 focuses on the details of the proposed approach. Section 5 summarizes the results of experiments. Finally, section 6 concludes the paper. The project page is available at: \url{https://patrykni.github.io/UnitySplat2Data/}.

\section{Related Works}
 
Large annotated datasets remain a critical bottleneck for deep learning in vision and robotics. Synthetic data generation offers a scalable solution, and early efforts relied on 3D rendering engines such as Blender or Unity, or simulation platforms like Falcon~\cite{bogdanoff_duality}. While powerful, these methods require significant manual effort to create high-quality 3D assets and accurately model environments to bridge the "domain--gap" to real-world conditions. Simpler strategies like ``Cut-and-Paste''~\cite{Ghiasi2020SimpleCI} expand datasets by compositing object crops onto backgrounds, but suffer from limited viewpoint diversity and contextual artifacts.  

The IJCV 2018 editorial~\cite{gaidon2018reasonable} articulated the trade-offs between realism, scalability, and the simulation-to-reality gap. Subsequent works confirmed the value of synthetic data: \cite{sakaridis2018foggy} showed foggy-scene synthesis improves detection under adverse conditions; \cite{shooter2024sydogvideo} presented SyDog-Video, demonstrating gains in temporal pose estimation from synthetic training; and \cite{souza2024naft} proposed SynthStab, coupling a new dataset with a RAFT-based stabilizer. In recent paper \cite{biswas2025compositionality}, the leveraged multifaceted synthetic corpora to enhance compositional object detection was proposed. Procedural pipelines such as BlenderProc automate photorealistic rendering with labels~\cite{denninger2023blenderproc2}, while domain randomization~\cite{tobin2017domainrandomization} and benchmarks like Syn2Real~\cite{peng2018syn2real} highlight both successes and remaining gaps in transfer.

Neural radiance fields (NeRF)~\cite{mildenhall2020nerf} inspired interest in learned scene representations, but rendering costs limit scalability. 3D Gaussian Splatting (3DGS)~\cite{kerbl3Dgaussians} overcomes this with real-time, high-fidelity novel view synthesis, making it attractive for dataset pipelines. Several concurrent works explore its potential. 

The advent of 3D Gaussian Splatting (3DGS) has introduced a new paradigm for synthetic data generation, enabling the creation of photorealistic representations directly from a set of posed multi-view images. This has inspired numerous works to develop data generation pipelines leveraging this technology. Moreover, the broader capture-reconstruct-generate architecture is increasingly recognized as an effective framework in robotics, a trend highlighted in recent surveys~\cite{irshad2024neural} and demonstrated by specific applications like SparseGrasp, which employs 3DGS for scene reconstruction to facilitate robotic grasping~\cite{yu2024sparsegrasp}. 

The PEGASUS system~\cite{pegasus_meyer} presents an architecture that is
conceptually the closest to ours. It separately reconstructs both environments and objects using 3DGS and leverages a physics engine for natural object placement. While PEGASUS provides a robust framework for scene composition, our method introduces a significant enhancement in visual fidelity through a dedicated post-processing pipeline for shadow and light integration. Our approach generates a separate shadow pass from object meshes, which is then algorithmically processed—involving steps like Gaussian blur and non-linear value mapping—before being composited with the splat render to create geometrically-correct, soft shadows. This directly addresses a known limitation of the PEGASUS system, which lacks realistic shadow rendering, thereby further improving visual fidelity.

Other works integrate 3DGS into different frameworks. The "Cut-and-Splat" method ~\cite{cut_and_splat_vanhele} composites 3DGS objects onto 2D background images,
ingeniously using monocular depth estimation to find plausible support surfaces for placement. However, their approach does not simulate lighting or shadows, relying instead on appearance augmentation to create visual variance. A different strategy is seen in the work~\cite{deogan_robot_soccer}, where import 3DGS objects into a traditionally modeled 3D environment within Unreal Engine. While our method also leverages a game engine (Unity), its role is fundamentally different. It relies entirely on Unreal Engine’s built-in renderer for their final image output. In contrast, our approach uses a hybrid rendering technique. We utilize Unity’s lighting system to generate a separate shadow pass by rendering the object meshes. This shadow pass is then composited with a render of the photorealistic splats. This final compositing step, which combines elements from different rendering passes, allows us to create highly realistic images with plausible shadows,
giving us unique control over the final appearance. 
 
A key limitation in many synthetic pipelines is the treatment of shadows and global illumination, which strongly affect realism. Traditional techniques addressed seamless compositing~\cite{chuang2003shadowmatting}, but recent methods improve controllability: \cite{sheng2022pixelheightshadow} generate shadows from pixel height maps, and \cite{valenca2023shadowharmonization} harmonized inserted shadows with scene illumination However, most 3DGS-based systems do not yet integrate physically plausible shadows. PEGASUS~\cite{pegasus_meyer} and Cut-and-Splat~\cite{cut_and_splat_vanhele} lack shadow simulation, while~\cite{deogan_robot_soccer} rely on Unreal’s built-in renderer. In contrast, our approach introduces a hybrid rendering strategy: we isolate splats for photorealism and combine them with a separate, physics-based shadow pass generated in Unity. This compositing step yields controllable, physically grounded shadows that improve both realism and detector robustness.
  
Orthogonal to rendering pipelines, some research improves the fidelity of source assets. For instance, Moghadam et al.~\cite{MoghadamDefectDetection} demonstrated a method for simulating manufacturing flaws by directly modifying 3D models to include realistic cracks and imperfections. Such methods complement advances in rendering by ensuring that the input assets themselves better approximate real-world variability.

Related to the above articles, our work is uniquely positioned at the intersection of these emerging techniques. Our pipeline addresses key realism gaps found in related works by combining a fully 3DGS-based representation for both objects and environments with a novel, explicit shadow simulation layer. This hybrid approach, orchestrated within the Unity engine, aims to generate highly realistic, physically plausible, and automatically annotated data for training robust robotic perception models.

\section{Preliminaries}

\subsection{Object detection and segmentation}

Object detection is a fundamental computer vision task that involves identifying and localizing objects within an image, typically by drawing bounding boxes around them. Instance segmentation is a more advanced task that goes a step further by predicting a pixel-level mask for each detected object instance, providing a much more detailed understanding of the object's shape and boundaries~\cite{liu2020deepSurvey,ren2024objectsegmentation}. In our experiments, we have chosen the YOLO (You Only Look Once) family of models, particularly the recent versions developed by Ultralytics, which have become a de facto standard for real-time object detection due to their high speed and accuracy~\cite{kotthapalli2025yolov1v11, ramos2025decadeyolo}. A key feature of this architecture is its scalability; it is offered in several variants of increasing size and complexity, typically denoted as nano (n), small (s), medium (m), large (l), and extra-large (x). This range provides a direct trade-off between inference speed and detection accuracy, allowing practitioners to select the optimal model for their specific constraints, whether for real-time applications on resource-constrained devices or for achieving maximum performance on high-end hardware. 
Notably, the training pipeline for YOLO's detection mode can accept data in a segmentation format, which consists of a text file for each image~\cite{ultralytics_yolo11}. In this format, each line represents a single object instance and follows the structure: `<class\_index> <x1> <y1> <x2> <y2> ...`. The `class\_index` is an integer identifying the object's class, and the subsequent pairs of numbers are the coordinates of the polygon vertices that define the object's segmentation mask. A key aspect of this format is that all coordinates are normalized to a range of [0, 1] by dividing them by the image's width and height, respectively. Our proposed method leverages this capability by generating these precise, normalized segmentation coordinates automatically.

\subsection{Evaluation Metrics}

The evaluation of object detection and segmentation models hinges on classifying each prediction against a ground-truth annotation. The primary mechanism for this is the Intersection over Union (IoU) score, which quantifies the degree of overlap between the predicted and actual object boundaries. Based on a predefined IoU threshold, a prediction is classified as a True Positive (TP) if it correctly identifies an object, a False Positive (FP) if it is an incorrect detection, and a False Negative (FN) is noted for any ground-truth object the model fails to find.

From these counts, two fundamental metrics are derived: Precision, which measures the accuracy of the predictions ($TP / (TP + FP)$), and Recall, which measures the model's ability to find all relevant objects ($TP / (TP + FN)$). A trade-off typically exists between these two; improving recall by detecting more objects can often lower precision by introducing more errors. This relationship is captured by the Precision-Recall (P-R) curve. The Average Precision (AP) provides a single, comprehensive score for a class by calculating the area under this P-R curve, summarizing the model's performance across all confidence thresholds.

The final metric, mean Average Precision (mAP), is the average of AP scores across all object classes. In our experiments, we use two standard variants of this metric. The mAP50 score is calculated using a lenient IoU threshold of 50\%, rewarding models for general object localization. The mAP50-95 score is a stricter evaluation, averaging mAP scores over ten IoU thresholds from 50\% to 95\%, demanding high precision in both location and shape. We report both Box mAP and Mask mAP, which apply these principles to bounding boxes and pixel-level masks, respectively. Reporting both scores thus provides a more comprehensive assessment, distinguishing between the model's ability to correctly identify objects and its capacity for their precise localization.

\subsection{Gaussian splatting}

3D Gaussian Splatting (3DGS) is a rasterization-based method for real-time rendering of photorealistic virtual views from a set of input images -- novel views synthesis~\cite{kerbl3Dgaussians}. In contrast to Neural Radiance Fields (NeRFs), which rely on continuous, implicit scene representations and computationally expensive volumetric ray-marching~\cite{mildenhall2020nerf}, 3DGS utilizes an explicit and unstructured representation based on a collection of 3D Gaussians. This approach achieves state-of-the-art visual quality while maintaining competitive training times and enabling real-time rendering at high resolutions.

The core primitive of the method is a 3D Gaussian, which is defined by several key attributes: a 3D position (mean) $\mu$, an anisotropic 3D covariance matrix $\Sigma$ (representing its scale and rotation), a color (represented by Spherical Harmonics (SH) coefficients to model view-dependent effects), and an opacity value $\alpha$. This explicit representation avoids the need for costly neural network evaluations during rendering, which is a primary reason for its high performance~\cite{kerbl3Dgaussians}.

\begin{figure}[h!]
    \centering
    \includegraphics[width=\textwidth]{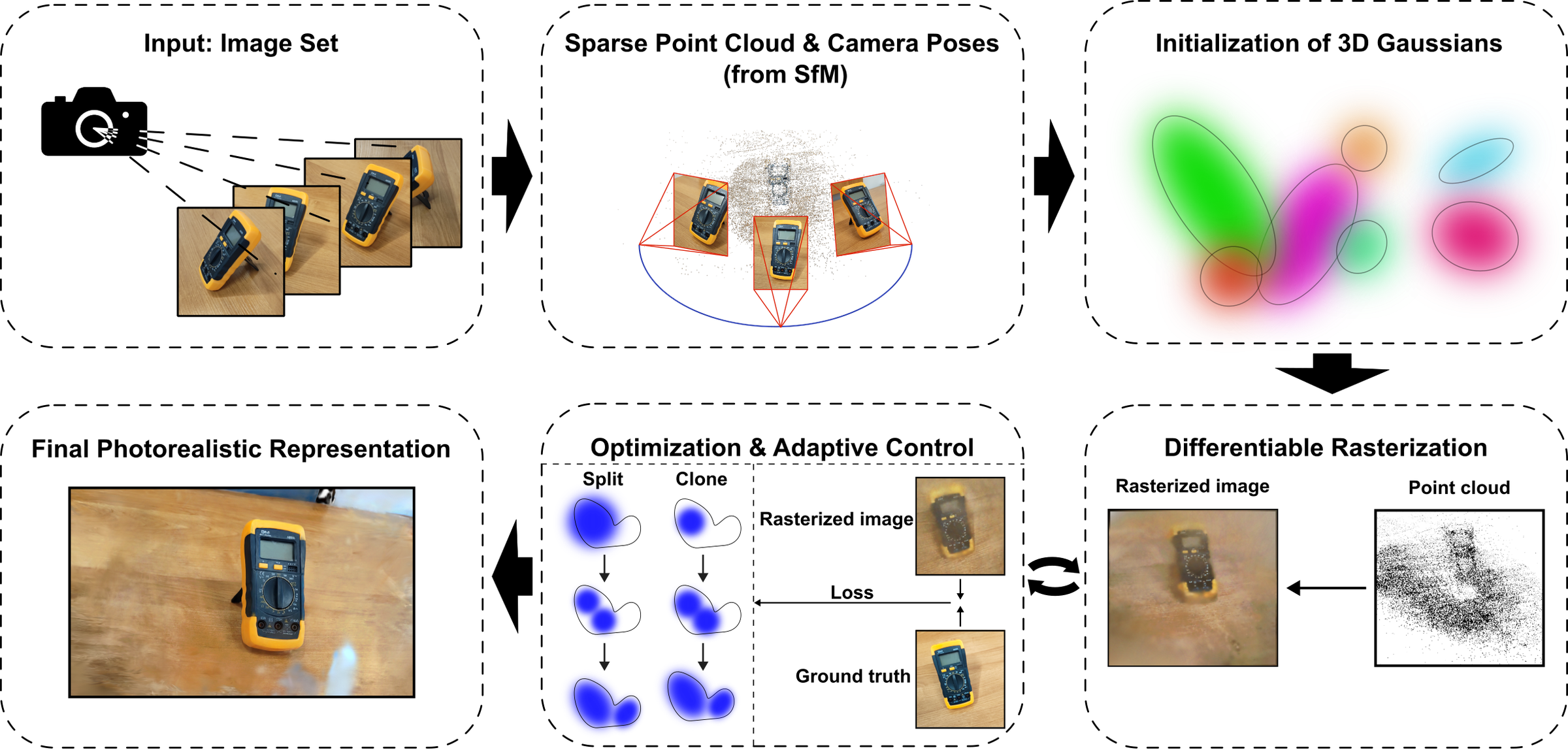}
    \caption{The 3D Gaussian Splatting pipeline.}
    \label{fig:gs_pipeline}
\end{figure}

The process, illustrated in Figure \ref{fig:gs_pipeline}, begins with data acquisition, as depicted in the "Input: Image Set" block of the diagram. This initial step involves capturing photographs of a static scene from various viewpoints. To enhance geometric accuracy, this input dataset can be augmented with depth maps corresponding to the input images. Subsequently, in the "Sparse Point Cloud \& Camera Poses" stage, Structure-from-Motion (SfM) software, such as \textit{COLMAP}, analyzes the input images~\cite{schonberger2016mvs} to yield the precise 3D position and orientation of the camera for each photograph and a sparse 3D point cloud that provides a preliminary geometric scaffold of the scene.

The third stage is the "Initialization of 3D Gaussians". In this step, the sparse point cloud generated by SfM is transformed into an initial set of 3D Gaussians. Each point from the cloud is converted into a single Gaussian primitive. The initial 3D position and the base color of each Gaussian are directly inherited from the corresponding point's attributes in the SfM output. This provides a coarse but geometrically aligned starting point for the subsequent optimization phase.

The core of the method is an iterative optimization loop, represented by the interconnected "Differentiable Rasterization" and "Optimization \& Adaptive Control" blocks of a diagram. The loop begins with rasterization, a process that projects all 3D Gaussians onto a 2D image plane to create a rendered image. This rasterizer is \textit{differentiable}, with is a crucial property allowing the calculation of gradients for every pixel with respect to the parameters of the Gaussians that influenced it. The rendered image is then compared against the corresponding ground-truth photograph, and a loss value is computed. If depth maps were provided as input, an additional depth loss term is calculated by comparing the rendered depth with the input depth map. This depth-regularized loss provides a strong geometric guide, helping to reduce artifacts and improve the accuracy of the reconstruction.

This loss value then drives the "Optimization \& Adaptive Control" stage. Using the gradients obtained from the differentiable rasterization, an optimization algorithm updates all the parameters of the 3D Gaussians to minimize the loss. This includes not only their 3D position $\mu$, but also their anisotropic covariance $\Sigma$ (which defines their 3D scale and rotation), their opacity $\alpha$, and their Spherical Harmonics (SH) coefficients, which model complex, view-dependent color effects. Concurrently, an adaptive control mechanism refines the geometry. As shown on the left of the optimization block, Gaussians in poorly represented areas are either cloned to add detail (under-reconstruction) or split into smaller Gaussians to refine complex surfaces (over-reconstruction). This entire loop of rasterization, loss calculation, and optimization is repeated thousands of times.

Once the optimization process converges and the loss is minimized, the final set of optimized 3D Gaussians constitutes the "Final Photorealistic Representation". This output is essentially a dense, highly detailed point cloud where each point contains rich information about position, shape, color, and opacity. This final representation can be saved and loaded into compatible viewers that utilize the same fast rasterization technique to enable real-time, high-quality navigation and rendering of novel views of the captured scene.

\section{Proposed method}

\begin{figure}[h!]
    \centering
    \includegraphics[width=\textwidth]{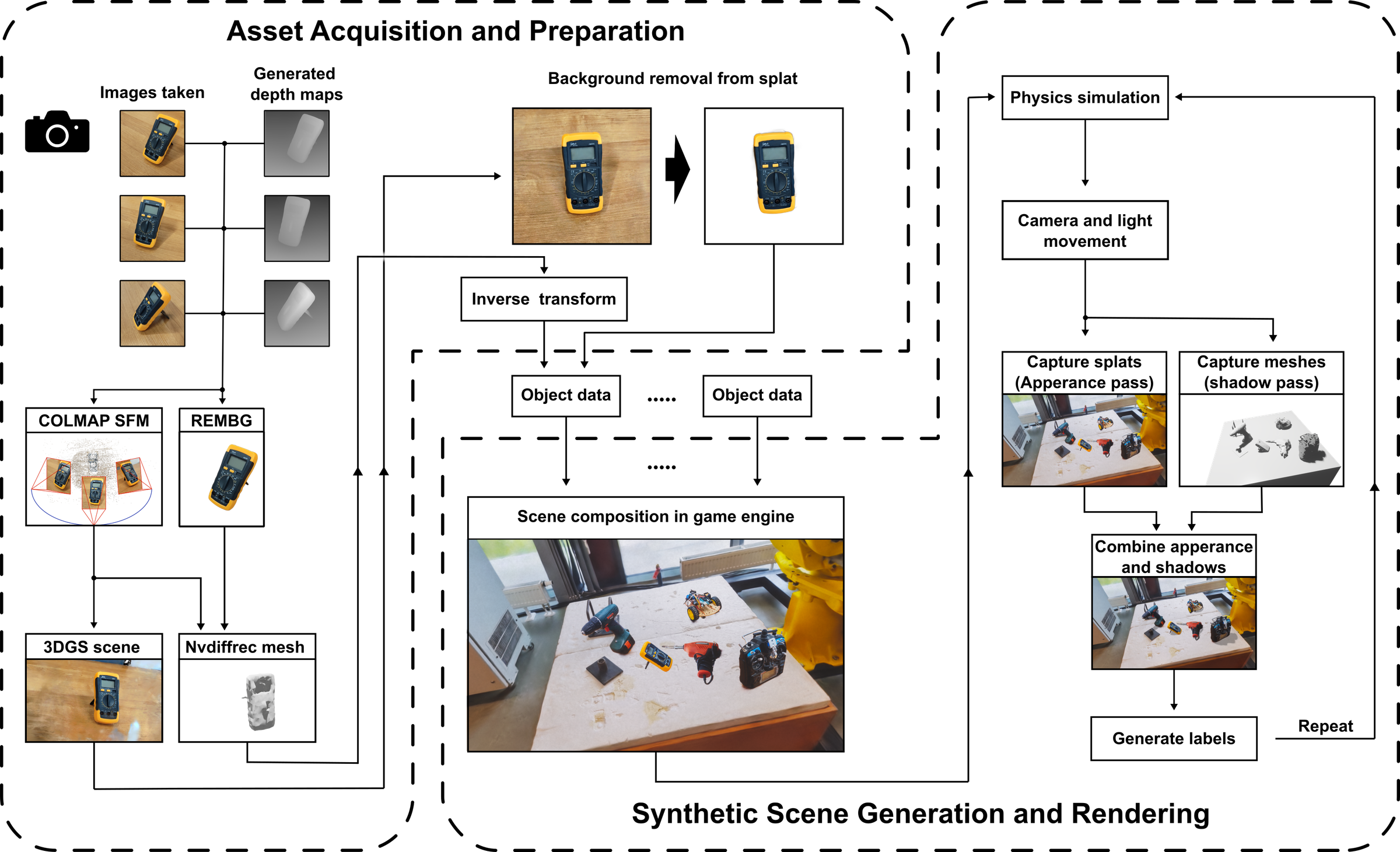}
    \caption{Diagram of the proposed method for generating synthetic datasets. }
    \label{fig:proposed_pipeline}
\end{figure}

Our proposed pipeline is an end-to-end system designed to transform a series of photographs into a large-scale, accurately labeled dataset suitable for training deep-learning-based object-detection and segmentation models. The entire process, illustrated in Figure \ref{fig:proposed_pipeline}, is divided into two main stages: "Asset Acquisition and Preparation", and "Synthetic Scene Generation and Rendering". The process begins with image acquisition, followed by the creation of 3D assets (splats and meshes). Scene composition and physics simulation take place in the Unity engine. The final image is created through hybrid rendering of splats and shadows, and labels are generated automatically, creating a complete dataset for training object detection and segmentation models, such as YOLO.

\subsection{Asset Acquisition and Preparation}
The first stage involves creating high-fidelity digital representations of the visions system environment and the objects of interest. The quality of these assets is fundamentally dependent on the initial set of captured images, the number and nature of which are dictated by the requirements of the Structure-from-Motion (SfM) algorithm. Successful SfM reconstruction is contingent upon providing a sufficient number of images with significant visual overlap, allowing the algorithm to match corresponding features and accurately determine camera poses. The acquisition technique involves capturing images from a multitude of viewpoints to ensure complete coverage. The required number of images depends on the subject's size and complexity; a large, detailed environment necessitates more images than a single object. While extracting frames from video is an option, we found that individual photos yield higher-quality results, avoiding issues like motion blur that can degrade reconstruction.

A counterintuitively critical aspect of this stage, particularly for individual objects, is the nature of the background. A feature-rich, textured background provides a wealth of stable keypoints essential for robust camera pose estimation, especially when the object itself is smooth. The ideal background should have a non-repetitive, high-frequency texture.

Our specific process, as detailed in the left panel of Figure \ref{fig:proposed_pipeline}, begins by capturing a comprehensive set of images of the target environment. These images, along with corresponding depth maps generated using the \textit{Depth Anything v2} model~\cite{depth_anything_v2_2024}, are processed to create a photorealistic 3D Gaussian Splat of the entire static scene. For each individual object, we perform several parallel tasks. First, camera poses are determined from the original, unmodified images using a Structure-from-Motion implementation (\textit{COLMAP}~\cite{schonberger2016mvs}). These poses are a prerequisite for generating an initial 3DGS of the object with its feature-rich background intact. Concurrently, clean object masks are created by programmatically removing the background from the same image set, for which we utilized the \textit{REMBG} library~\cite{rembg}. Finally, a precise mesh of the object is generated using these background-free images and the previously calculated camera poses. This is achieved through a differentiable inverse graphics approach, implemented with the \textit{nvdiffrec} tool~\cite{munkberg2021nvdiffrec}. Finally, to isolate the object’s splat representation, we perform the 'Background removal from splat' step shown on the diagram by projecting the generated masks into the 3D scene and filtering the initial Gaussian point cloud. A crucial final step is to align the generated mesh with its corresponding splat representation. The \textit{nvdiffrec} tool automatically applies a transformation to the input camera poses to center and scale the object, which simplifies the mesh generation process. However, this results in a coordinate system mismatch between the output mesh and the original Gaussian Splat. To rectify this, we apply the 'Inverse transform', ensuring both assets are perfectly co-located. The outcome of this process is a database of asset pairs, where each object is represented by both its optimized splat representation and its corresponding aligned mesh.

\subsection{Synthetic Scene Generation and Rendering} 
With all assets prepared, we move to the "Synthetic Scene Generation and Rendering" stage shown on the right of the diagram. We use the Unity engine, enhanced with the \textit{UnityGaussianSplatting} library~\cite{UnityGaussianSplatting_github}, to generate the final dataset. A new scene is created for each environment, where we load the environment splat and augment it with simple 3D shapes. These shapes serve multiple purposes: some are visible elements with colliders, such as a tabletop, while others are completely invisible colliders that form an enclosure around the workspace, preventing objects from falling off during the physics simulation. The previously generated assets for each object—the photorealistic Gaussian Splat and its corresponding collision mesh from nvdiffrec—are then loaded into the scene. A script ensures that each splat's transform matches its mesh after physics simulation. For each generation series, object meshes are randomly positioned above the table and dropped, allowing the 'Physics simulation' step to create a natural resting pose. The scene's camera and light sources are also animated along predefined paths to ensure variety, as indicated by the "Camera and light movement" block.

\begin{figure}[h!]
    \centering
    \includegraphics[width=\textwidth]{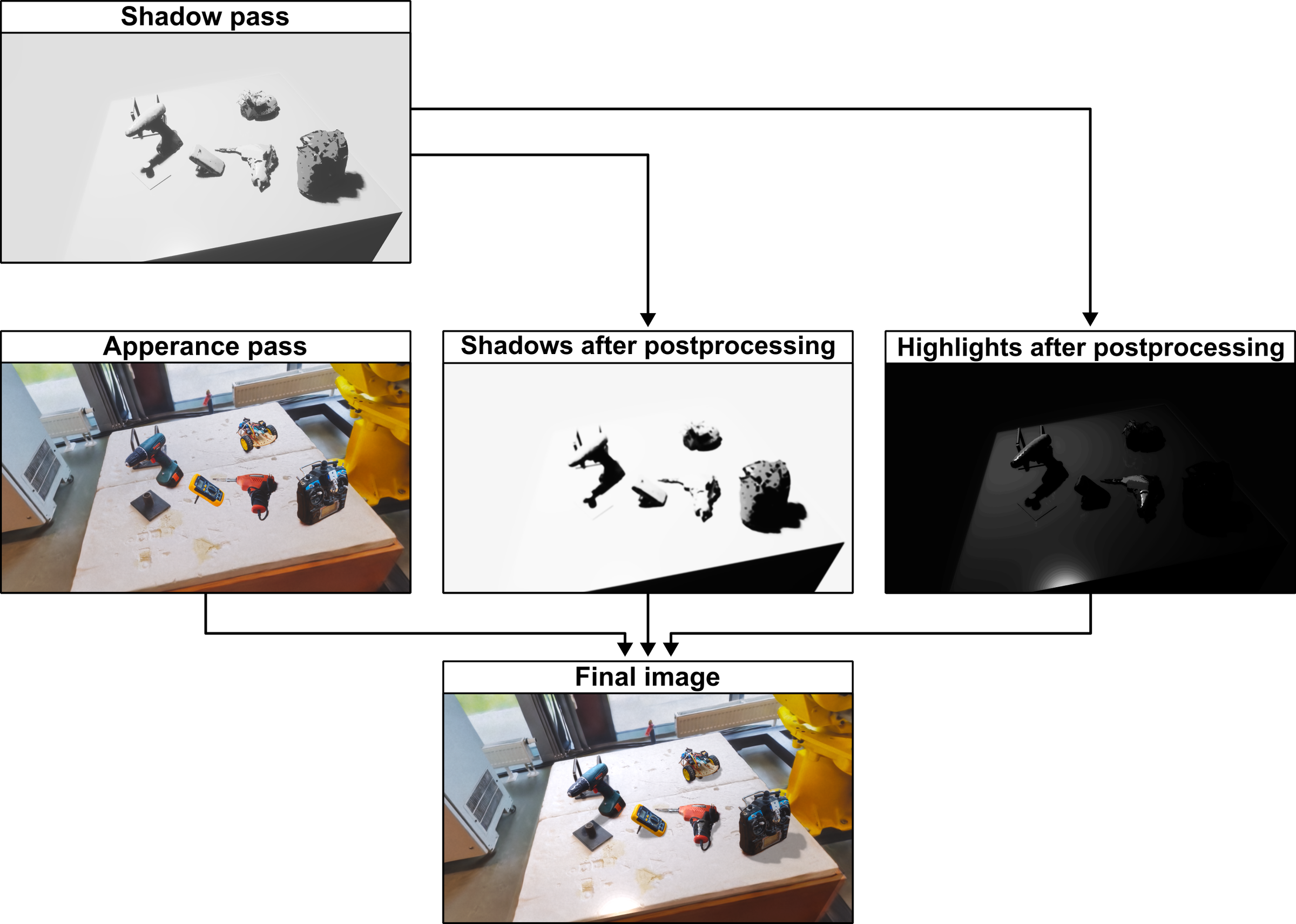}
    \caption{Hybrid compositing pipeline}
    \label{fig:hybrid_rendering}
\end{figure}

A key innovation of our method is a two-pass hybrid rendering approach, shown in Figure \ref{fig:hybrid_rendering}. For each frame, a photorealistic render of the Gaussian Splats is generated (the "Appearance pass") alongside a separate pass from the simple object meshes (the "Shadow pass"). Importantly, this pass includes meshes of both the objects and key scene elements (e.g., the tabletop), allowing the system to naturally handle complex interactions such as shadows cast by one object onto another and by the environment onto the objects. This shadow map undergoes a multi-stage algorithmic process: it is first normalized and softened using a Gaussian blur, then its tonal transitions are refined with a sigmoid curve. The final composition multiplicatively applies these processed shadows to the splat render. Furthermore, the same map is used to create a highlight mask, which is additively blended with the image to simulate light reflections. This advanced compositing technique combines the photorealism of Gaussian Splatting with physically plausible shadows and lighting. To further increase data variance, random post-processing effects such as adjustments to hue, exposure, and noise are applied.

\subsection{Automated Label Generation}
\begin{figure}[h!]
    \centering
    \includegraphics[width=\textwidth]{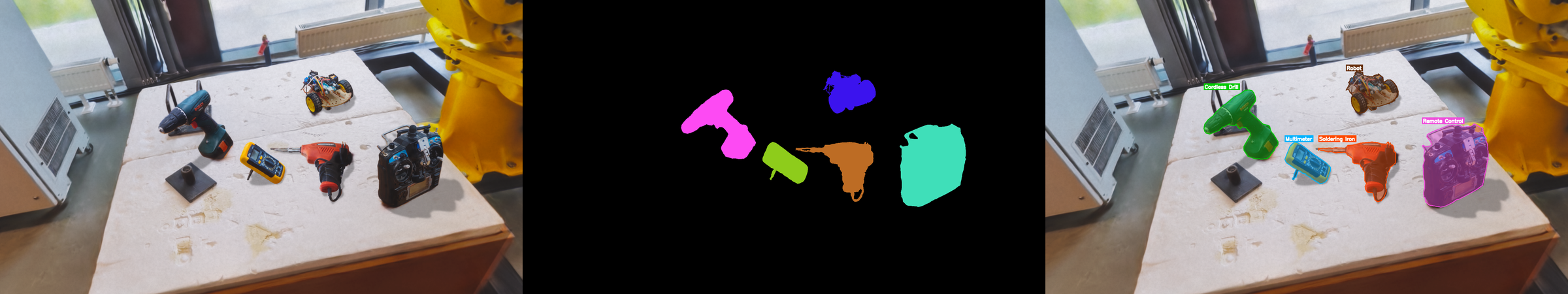}
    \caption{The automated labeling process. From left to right: the final rendered image, the generated instance segmentation masks (ID map), and the final image with bounding boxes and segmentation contours applied.}
    \label{fig:labeling_process}
\end{figure}

Leveraging the controlled environment of the game engine, we can generate pixel-perfect labels automatically. This process is visualized in Figure \ref{fig:labeling_process}. For each final image, we create a corresponding instance segmentation mask. This is achieved by rendering each object's mesh to an in-memory buffer with a unique solid color, respecting the render order to handle occlusions correctly. The resulting multi-colored "ID map" is then processed; by identifying pixels of a specific color, we extract the precise contour of each visible object part. Small, noisy mask fragments are filtered out. The final output for each generated scene is a pair: the composite image with shadows and a text file containing the class and segmentation coordinates in required data format e.g. the YOLO standard.

\section{Experiments}

\begin{figure}[h!]
    \centering
    \includegraphics[width=\textwidth]{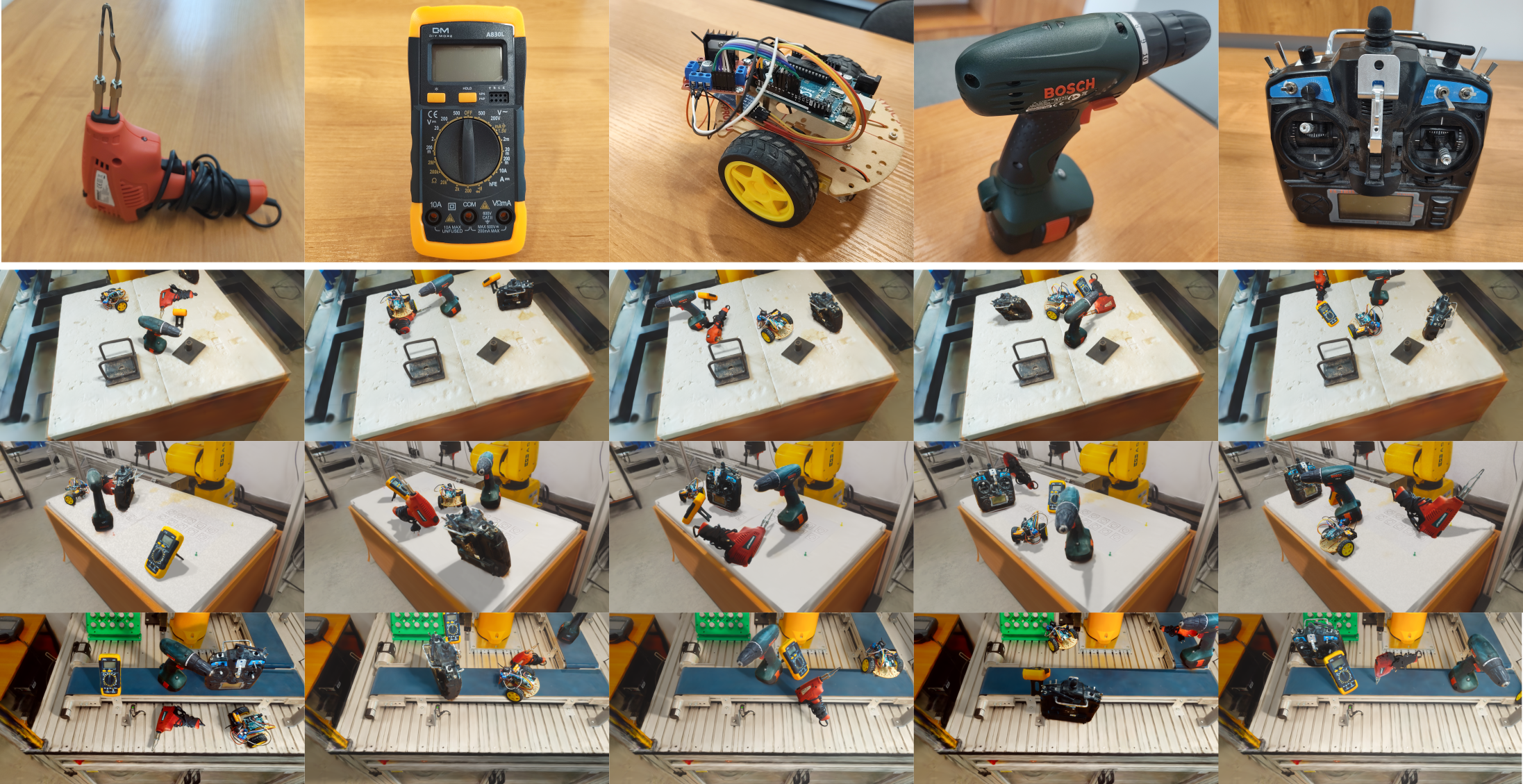}
    \caption{Objects used in the experiments (top row) and example scenes showing their example placement in the robotic workstations.}
    \label{fig:PhotosObjectsAndPlacement}
\end{figure}

Our proposed pipeline is primarily designed for robotics applications, specifically for perception systems operating within an industrial robot's workspace. To validate our approach in its intended context, all experiments were conducted within such a scenario. The setup involved various objects of interest being placed within the operational area of a robotic workstation, simulating a typical environment for tasks like robotic pick-and-place or assembly (Figure \ref{fig:PhotosObjectsAndPlacement}).

\begin{figure}[h!]
    \centering
    \includegraphics[width=0.7\textwidth]{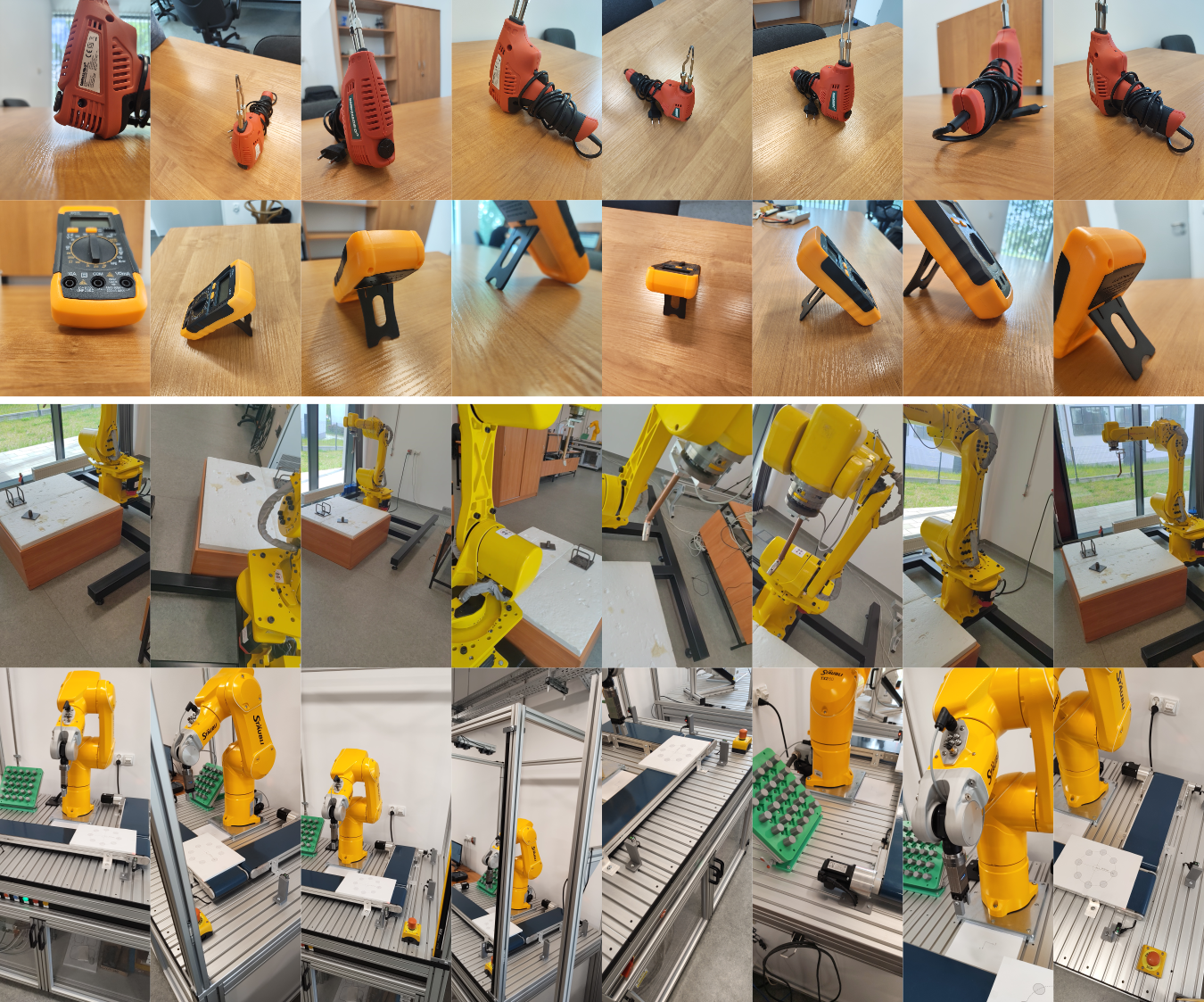}
    \caption{Example images from the data acquisition stage, showing the capture process for both a single object (top) and the environment (bottom).}
    \label{fig:example_photos}
\end{figure}

To evaluate our proposed data generation pipeline, we conducted a series of experiments. The first step involved applying the acquisition method described in Section 4.1 to capture the source images for our 3D assets. Through empirical evaluation, we found that approximately 300 images were sufficient for a complex environment (a robotic workstation), while around 60 images yielded a high-quality reconstruction for a single object. These values represent a practical balance, ensuring sufficient feature coverage for a high-quality SfM reconstruction while avoiding the diminishing returns associated with a significantly larger image set. Examples of the captured images are shown in Figure \ref{fig:example_photos}.

With the 3D assets prepared, we proceeded to generate our synthetic dataset. To evaluate its effectiveness, we designed a series of experiments to compare the performance of a YOLO11 model~\cite{ultralytics_yolo11} trained on different data configurations. Specifically, we trained separate models on a purely real dataset, our purely synthetic dataset, and a hybrid combination of both. All models were then benchmarked on a consistent, held-out test set of real images. The primary goal was to quantify the performance of a model trained on our synthetic data against one trained on real-world data and to assess the benefits of a hybrid training strategy.

\subsection{Datasets}
We prepared several distinct datasets for our experiments to thoroughly test different training strategies. All models were ultimately evaluated on the same real test set to ensure a fair and consistent comparison.

\begin{figure}[h!]
    \centering
    \includegraphics[width=\textwidth]{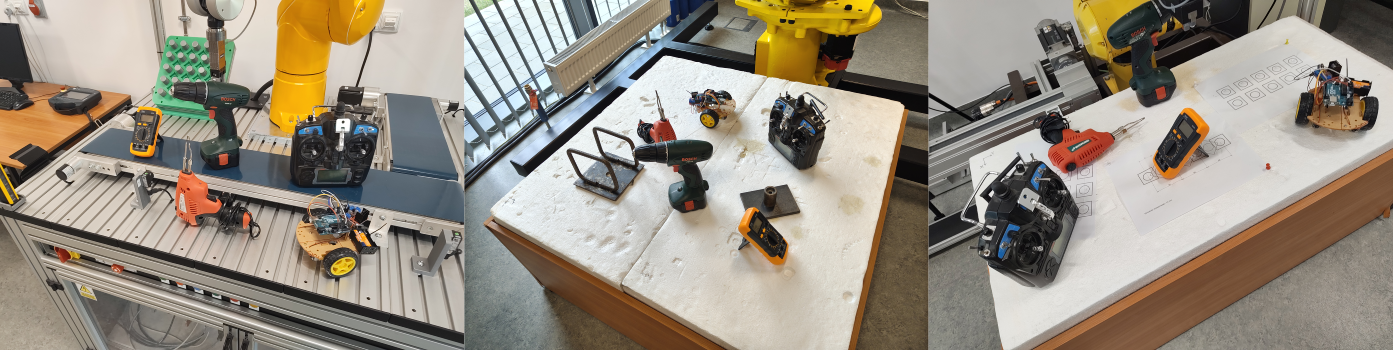}
    \caption{Example images from the Real Training Set.}
    \label{fig:real_train}
\end{figure}

\begin{figure}[h!]
    \centering
    \includegraphics[width=\textwidth]{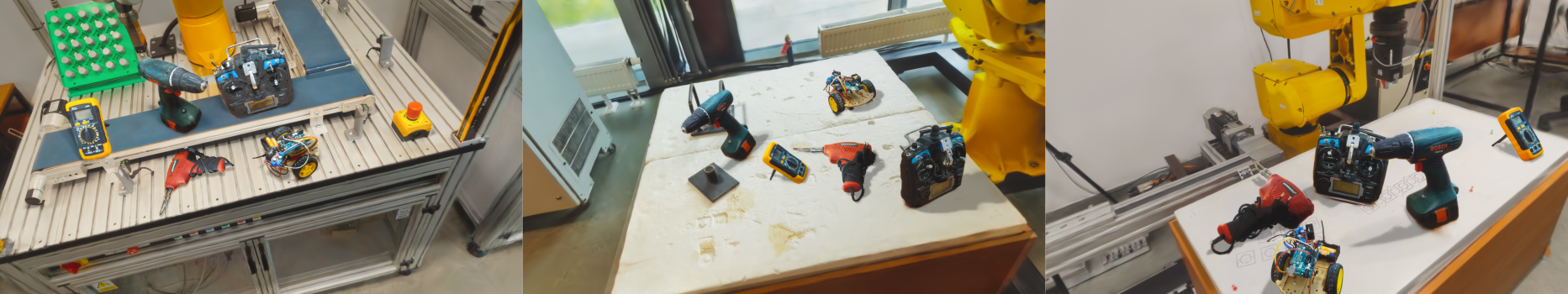}
    \caption{Example images from the Synthetic Training Set.}
    \label{fig:synth_train}
\end{figure}

\begin{figure}[h!]
    \centering
    \includegraphics[width=\textwidth]{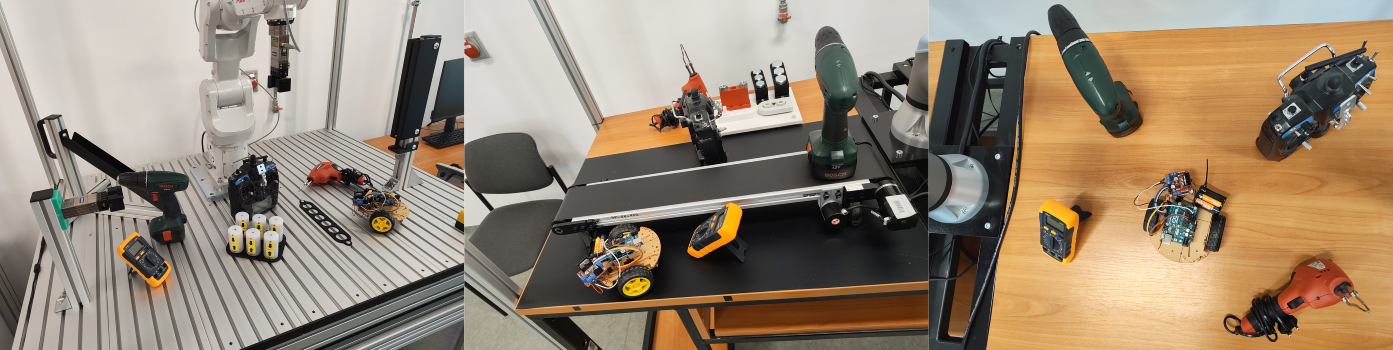}
    \caption{Example images from the Real Test Set.}
    \label{fig:real_test}
\end{figure}

\begin{itemize}
    \item \textbf{Real Training Set:} This dataset consists of manually captured and annotated images from our robotic lab environments. To analyze the impact of dataset size, we created two versions: a smaller one with 25 training images and a larger one with 75 training images. Both versions share a validation set of 30 images. These images were captured across three different robotic workstations to provide a variety of background contexts. Example images are shown in Figure \ref{fig:real_train}.

    \item \textbf{Synthetic Training Set:} This dataset was generated entirely using our proposed pipeline and contains 900 training images and 300 validation images. The significant disparity in size compared to the real dataset highlights a key advantage of our method: the ability to rapidly generate large volumes of perfectly labeled data with minimal effort. Example images are shown in Figure \ref{fig:synth_train}.
    
    \item \textbf{Hybrid Training Set:} To evaluate the benefits of combining real and synthetic data, we created a hybrid set. It consists of the larger real training set (75 images) merged with the synthetic set (900 images), for a total of 975 training images. The validation set is similarly combined, resulting in 330 images (30 real + 300 synthetic).

    \item \textbf{Real Test Set:} This is a separate, held-out dataset of 93 real-world images, which was not used during the training of any model. To properly test the model's generalization capabilities, this data was collected from two different robotic workstations that were not part of the training set. One of these workstations included two distinct table setups, effectively creating three unique test environments. Example images are shown in Figure \ref{fig:real_test}.
\end{itemize}

\subsection{Experimental Setup and Results}
For the experiments, we trained several YOLO models of varying sizes on each of the prepared training sets. A batch size of 8 was used for all training runs. This value was selected as the maximum feasible size constrained by the available GPU memory for our largest model variant; it was kept consistent across all models to ensure a fair and controlled comparison. All models were trained for 150 epochs, a duration determined empirically to provide a balance between achieving model convergence and avoiding significant overfitting, while also maintaining computational efficiency. Other hyperparameters were kept at their default values. The performance of each trained model was then evaluated on the real test set using the mean Average Precision (mAP) metric, focusing on the mAP50 and mAP50-95 scores for both detection (Box mAP) and segmentation (Mask mAP).

\begin{figure}[h!]
\centering
\includegraphics[width=1\textwidth]{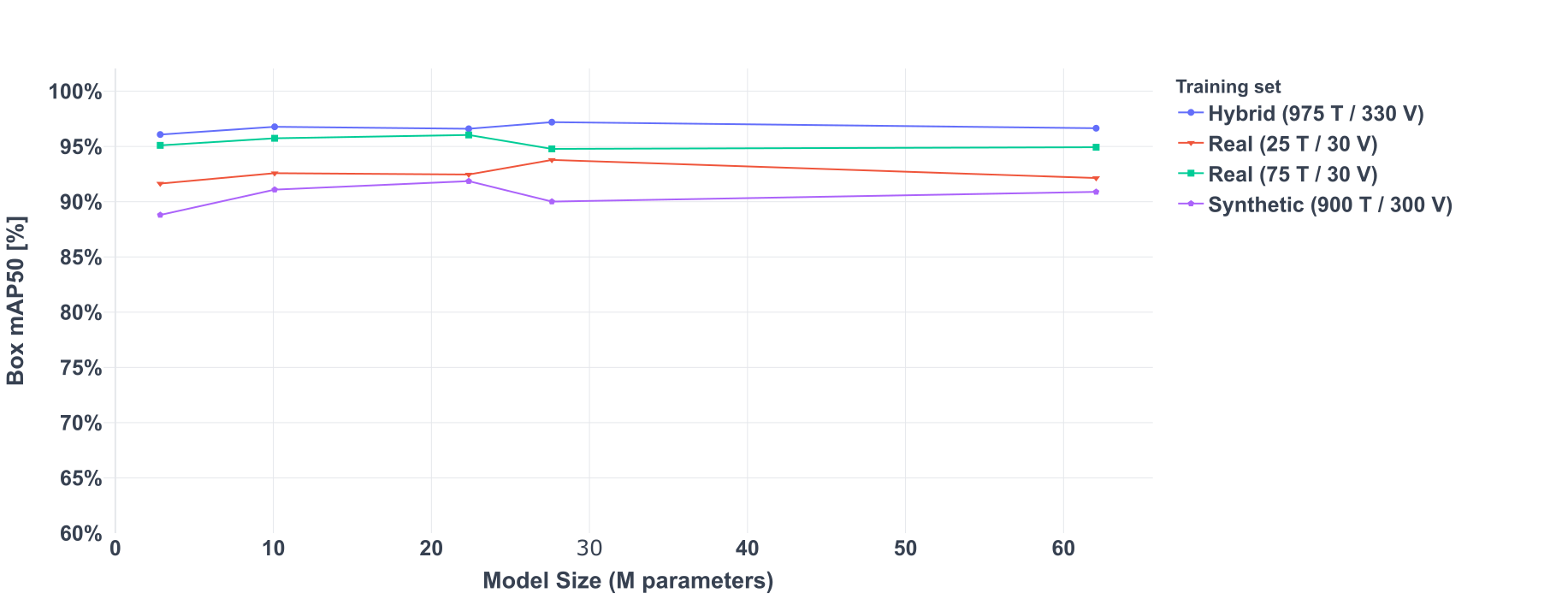}
\caption{Performance comparison for detection (Box mAP) at IoU threshold 0.5.}
\label{fig:map50_box_results}
\end{figure}

\begin{figure}[h!]
\centering
\includegraphics[width=1\textwidth]{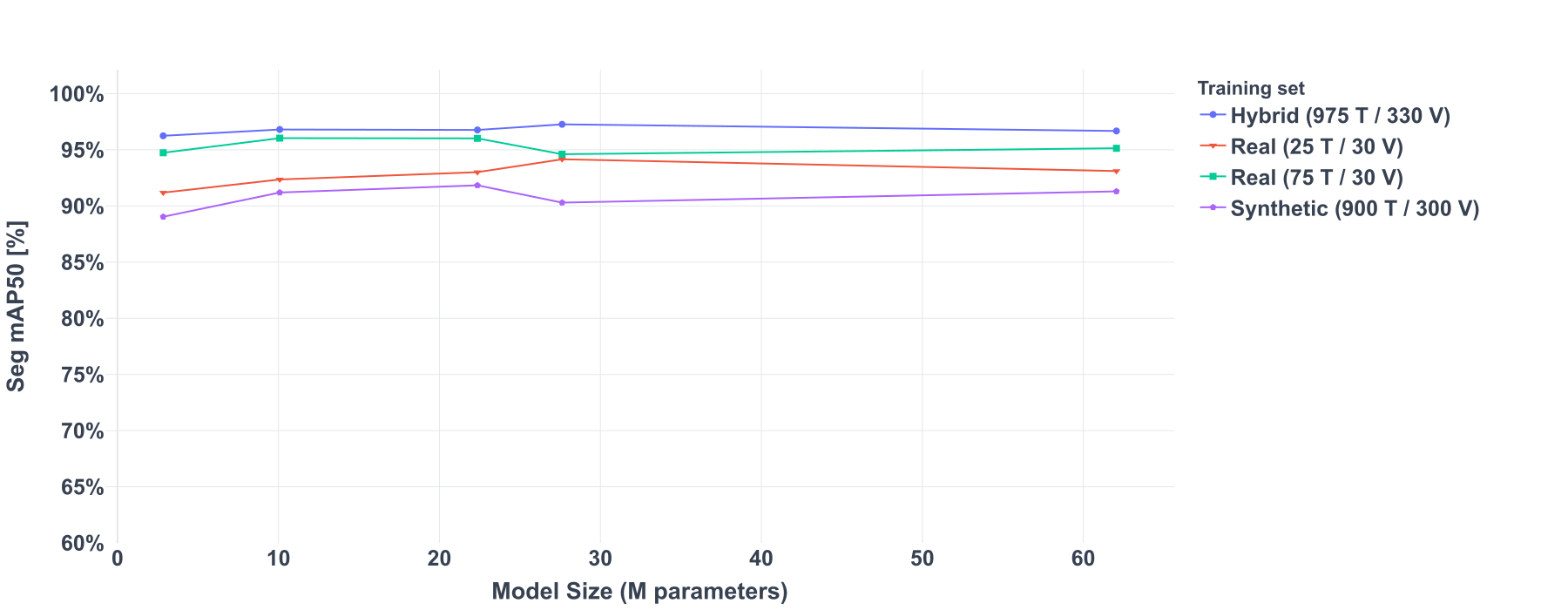}
\caption{Performance comparison for segmentation (Mask mAP) at IoU threshold 0.5.}
\label{fig:map50_mask_results}
\end{figure}

\begin{figure}[h!]
\centering
\includegraphics[width=1\textwidth]{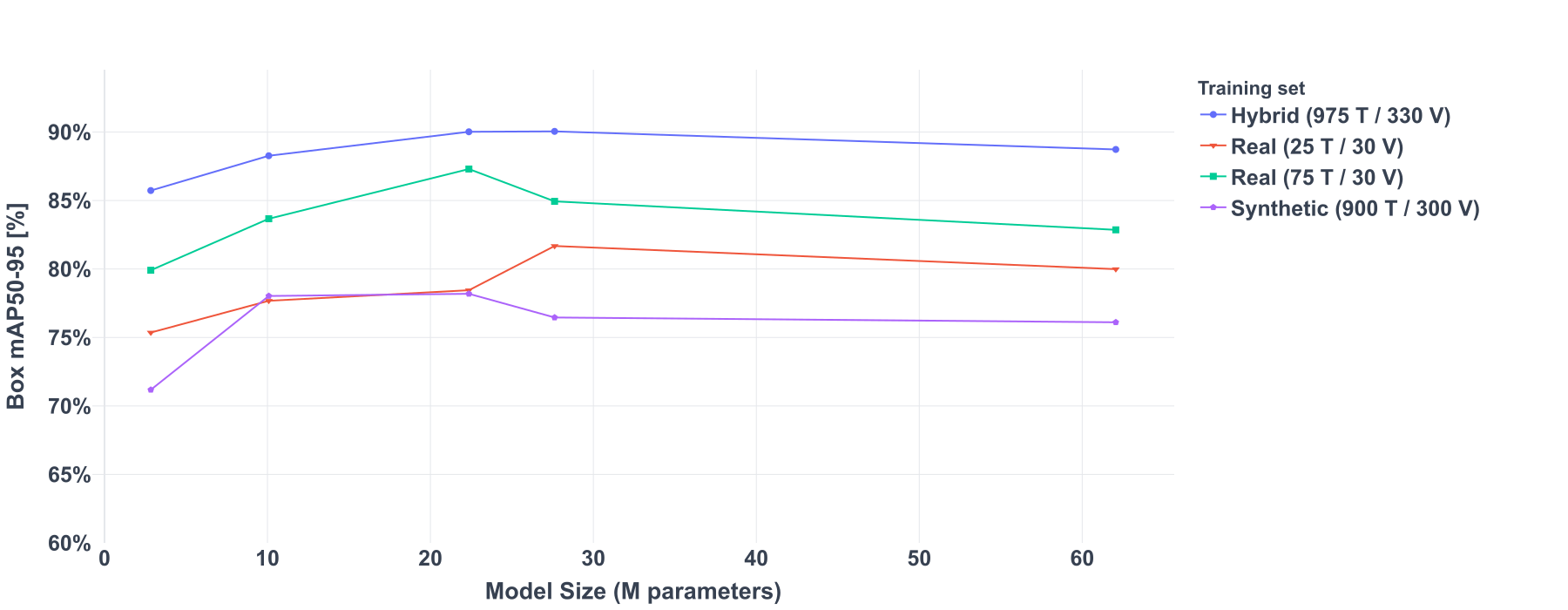}
\caption{Performance comparison for detection (Box mAP) averaged over IoU thresholds from 0.5 to 0.95.}
\label{fig:map5095_box_results}
\end{figure}

\begin{figure}[h!]
\centering
\includegraphics[width=1\textwidth]{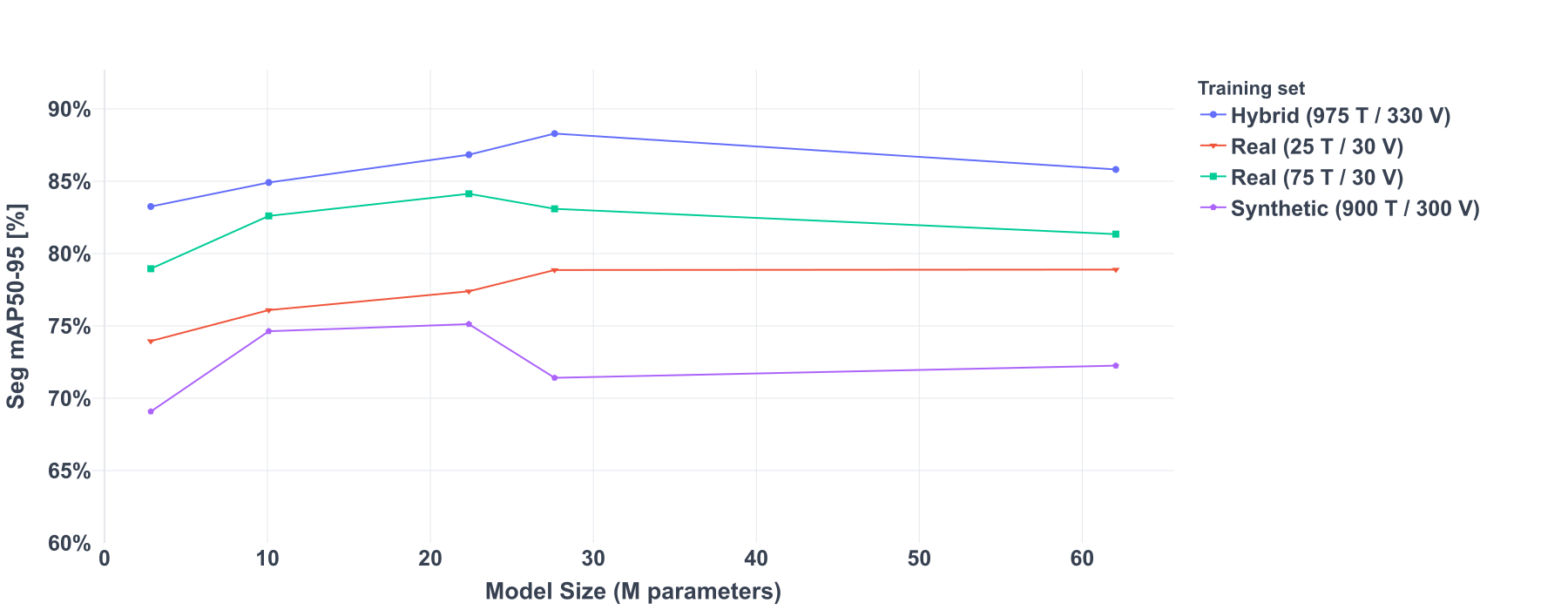}
\caption{Performance comparison for segmentation (Mask mAP) averaged over IoU thresholds from 0.5 to 0.95.}
\label{fig:map5095_mask_results}
\end{figure}

The detailed results are presented in Table \ref{tab:all_results}, with a graphical representation in Figures \ref{fig:map50_box_results} through \ref{fig:map5095_mask_results}. The Hybrid Training Set consistently achieves the highest performance across all model sizes and on all metrics. This strongly suggests that our synthetic data serves as a powerful augmentation to a smaller real dataset. The hybrid approach combines the two main advantages of the other sets: the high domain fidelity of the real images, which were captured with the same camera as the test set, and the vast variation in object poses, lighting, and backgrounds provided by the large synthetic dataset.

\begin{table}[h!]
\centering
\caption{Detailed performance metrics for all trained models on the test set. The best results for each metric and model size are shown in \textbf{bold}.}
\label{tab:all_results}
\small
\begin{tabular}{@{}lc*{4}{c}@{}}
\toprule
\textbf{Training Set} &
\textbf{\begin{tabular}[c]{@{}c@{}}Model size\\(M)\end{tabular}} &
\textbf{\begin{tabular}[c]{@{}c@{}}Box mAP50\\(\%)\end{tabular}} &
\textbf{\begin{tabular}[c]{@{}c@{}}Box mAP50-95\\(\%)\end{tabular}} &
\textbf{\begin{tabular}[c]{@{}c@{}}Mask mAP50\\(\%)\end{tabular}} &
\textbf{\begin{tabular}[c]{@{}c@{}}Mask mAP50-95\\(\%)\end{tabular}} \\
\midrule
\addlinespace[0.2em]
\multicolumn{6}{@{}l}{\textit{YOLO11n}} \\
\addlinespace[0.1em]
\midrule
Hybrid    & 2.84  & \textbf{96.08} & \textbf{85.73} & \textbf{96.25} & \textbf{83.25} \\
Real      & 2.84  & 91.64          & 75.36          & 91.19          & 73.95          \\
Synthetic & 2.84  & 88.80          & 71.18          & 89.04          & 69.08          \\
\midrule
\addlinespace[0.2em]
\multicolumn{6}{@{}l}{\textit{YOLO11s}} \\
\addlinespace[0.1em]
\midrule
Hybrid    & 9.43  & \textbf{98.99} & \textbf{88.69} & \textbf{98.41} & \textbf{86.55} \\
Real      & 9.43  & 98.23          & 85.24          & 97.02          & 82.51          \\
Synthetic & 9.43  & 94.27          & 77.12          & 91.62          & 74.23          \\
\midrule
\addlinespace[0.2em]
\multicolumn{6}{@{}l}{\textit{YOLO11m}} \\
\addlinespace[0.1em]
\midrule
Hybrid    & 20.06 & \textbf{98.59} & \textbf{87.73} & \textbf{97.85} & \textbf{85.24} \\
Real      & 20.06 & 97.89          & 85.01          & 96.54          & 82.09          \\
Synthetic & 20.06 & 94.88          & 79.25          & 92.55          & 76.88          \\
\midrule
\addlinespace[0.2em]
\multicolumn{6}{@{}l}{\textit{YOLO11l}} \\
\addlinespace[0.1em]
\midrule
Hybrid    & 25.31 & \textbf{98.84} & \textbf{88.05} & \textbf{98.11} & \textbf{85.73} \\
Real      & 25.31 & 98.15          & 85.33          & 96.98          & 82.47          \\
Synthetic & 25.31 & 95.01          & 79.58          & 92.98          & 77.01          \\
\midrule
\addlinespace[0.2em]
\multicolumn{6}{@{}l}{\textit{YOLO11x}} \\
\addlinespace[0.1em]
\midrule
Hybrid    & 56.88 & \textbf{98.77} & \textbf{87.93} & \textbf{98.05} & \textbf{85.58} \\
Real      & 56.88 & 98.02          & 85.19          & 96.88          & 82.13          \\
Synthetic & 56.88 & 95.23          & 80.12          & 93.12          & 77.54          \\
\bottomrule
\end{tabular}
\end{table}

The model trained on the larger Real Training Set (75 images) generally outperforms the model trained purely on synthetic data, especially on the stricter mAP50-95 metric. This is expected, as it minimizes the domain gap related to camera sensor characteristics. However, its performance is still notably lower than that of the hybrid model, highlighting the limitations of relying solely on a small number of real-world examples.

The Synthetic Training Set on its own achieves a respectable performance, validating the quality of our data generation pipeline. While it is outperformed by the other sets, it establishes a strong baseline that could be crucial in scenarios where collecting any real data is impractical. The performance gap can be attributed to the domain shift between the synthetic renders and the real test images.

The trends observed for both detection (Box mAP) and segmentation (Mask mAP) are remarkably similar, confirming that our pipeline generates accurate segmentation masks that effectively teach the model the precise shapes of the objects, leading to high-quality instance segmentation.

\section{Conclusions}

In this paper, we have presented a comprehensive, end-to-end pipeline for generating high-fidelity, automatically labeled synthetic datasets for training object detectors in robotic applications. Our method successfully leverages the photorealism of 3D Gaussian Splatting and introduces a novel hybrid rendering technique that composites physically-plausible shadows onto the scene. This approach effectively addresses the critical challenge of the domain gap between synthetic and real-world imagery, while the automated nature of the pipeline eliminates the time-consuming manual annotation bottleneck.

Our experiments systematically demonstrated the value of this approach. The results unequivocally show that the hybrid training strategy yields the best detection and segmentation performance across all tested models. This confirms that our synthetic data acts as a powerful augmentation, combining the sheer volume and variety of generated images with the high domain fidelity of real ones. It is important to note that the real training images were captured with the same camera as the test set, which gives them an inherent advantage. Consequently, the purely synthetic dataset, while achieving respectable results, was consistently outperformed by datasets containing real images. Nevertheless, its strong standalone performance validates our pipeline as a viable tool for bootstrapping perception systems in scenarios where collecting real data is impractical. Our findings confirm that combining a small set of real images with a large volume of our generated data is the optimal strategy for achieving robust and accurate models.

\section*{Statements and Declarations}

\textbf{Funding}
\newline
The authors did not receive support from any organization for the submitted work.

\vspace{\baselineskip}
\noindent\textbf{Competing Interests}
\newline
The authors have no relevant financial or non-financial interests to disclose.

\vspace{\baselineskip}
\noindent\textbf{Data, Material, and Code Availability}
\newline
The datasets and source code generated during the current study are available from the corresponding author on reasonable request for the purpose of peer review. Upon publication, the materials will be made publicly and permanently available via a GitHub Pages site.

\bibliography{datagensplat}

\end{document}